\documentclass{article}

\PassOptionsToPackage{numbers, compress}{natbib}

\usepackage[eandd, final]{neurips_2026}

\usepackage[utf8]{inputenc} 
\usepackage[T1]{fontenc}    
\usepackage{hyperref}       
\usepackage{url}            
\usepackage{booktabs}       
\usepackage{amsfonts}       
\usepackage{nicefrac}       
\usepackage{microtype}      
\usepackage{xcolor}         
\usepackage{graphicx}
\usepackage{pifont}%
\newcommand{\cmark}{\ding{51}}%
\newcommand{\xmark}{\ding{53}}%
\usepackage{multirow}
\usepackage{stmaryrd}
\usepackage{float}
\usepackage{amsmath}
\usepackage{enumitem}
\usepackage{bbm}
\usepackage{caption}
\usepackage{subcaption}

\makeatletter
\def\maketitlesupplementary{%
  \newpage
  \vbox{%
    \hsize\textwidth
    \linewidth\hsize
    \vskip 0.1in
    \@toptitlebar
    \centering
    {\LARGE\bf \@title\par}
    \vspace{0.5em}
    {\Large Supplementary Material\par}
    \@bottomtitlebar
  }
}
\makeatother

\title{Text-to-CAD Evaluation with \textsc{CADTests}}

\author{%
  Dimitrios Mallis \\
  {\small SnT, University of Luxembourg}
  \And
  Marco Wang \\
  {\small SnT, University of Luxembourg}
  \And
  Ahmet Serdar Karadeniz \\
  {\small SnT, University of Luxembourg}
  \AND
  Elisa Ricci \\
  {\small University of Trento} \\
  {\small Fondazione Bruno Kessler}
  \And
  Anis Kacem \\
  {\small SnT, University of Luxembourg}
  \And
  Djamila Aouada \\
  {\small SnT, University of Luxembourg}
}

\begin{document}

\maketitle

\begin{abstract}
  Text-to-CAD has emerged as an important task with the potential to substantially accelerate design workflows. Despite its significance, there has been surprisingly little work on Text-to-CAD evaluation, and assessing CAD model generation performance remains a considerable challenge. In this work, we introduce a new evaluation perspective for Text-to-CAD based on \textit{automated testing}. We propose \textsc{CADTestBench}, the first \textit{test-based} benchmark for Text-to-CAD, based on \textsc{CADTests}, executable software tests that verify whether a generated CAD model satisfies the geometric and topological requirements of the input prompt. Using \textsc{CADTestBench}, we conduct comprehensive benchmarking of recent Text-to-CAD methods and further demonstrate that \textsc{CADTests} can also guide CAD model generation, yielding simple baselines that surpass performance of current methods. 
   The \textsc{CADTestBench} dataset~\footnote{Hugging Face: \url{https://huggingface.co/datasets/dimitrismallis/CADTestBench}} and code~\footnote{GitHub: \url{https://github.com/dimitrismallis/CADTestBench}} are publicly availiable and be found also in the project page \url{https://dimitrismallis.github.io/cadtests/}.
\end{abstract}

\section{Introduction}
\label{sec:intro}

Computer-Aided Design (CAD) generative modeling has recently attracted significant attention from both academia and industry, particularly driven by the impressive emerging capabilities of Vision Language Models (VLMs). VLMs have the potential of reshaping traditional CAD workflows by extending the capabilities of conventional graphical design interfaces offered by widely used platforms such as FreeCAD~\cite{FreeCAD}, SolidWorks~\cite{solidworks} or Fusion 360~\cite{fusion360}. They enable generative design conditioned on text prompts~\cite{Yuan2025CADEditorAL, Guan2025CADCoderTG} as well as diverse input modalities, including images~\cite{mallis2025cad, Kolodiazhnyi2025cadrilleMC}, drawn sketches~\cite{seff2022vitruvion, Karadeniz2024PICASSOAF, Karadeniz2024DAVINCIAS}, point-clouds~\cite{wu2021deepcad,dupont2024transcad,Rukhovich2024CADRecodeRE, Kolodiazhnyi2025cadrilleMC} or 3D meshes~\cite{Karadeniz2025MiCADangeloFR, Karadeniz2026CADonatelloTR}. Among the various input modalities, recent work increasingly focuses on text-conditioned CAD generation~\cite{cadllama, Kolodiazhnyi2025cadrilleMC,khan2024text2cad,Yuan2025CADEditorAL,mallis2025cad,Badagabettu2024Query2CADGC,Idea2cad2025}. Language provides a natural and intuitive medium for human–computer interaction, which has led to the emergence of the Text-to-CAD research thread. 

Text-to-CAD research has primarily targeted improving generation performance, while rigorous evaluation has received far less attention. Current methods mostly rely on geometric similarity metrics~\cite{khan2024text2cad,alrashedy2024generating,Li2024CADTA,Yuan2025CADEditorAL,mallis2023sharp,Yavartanoo2024Text2CADTT,Wang2025TexttoCADGT,Guan2025CADCoderTG,Kolodiazhnyi2025cadrilleMC}, such as Chamfer Distance (CD), computed against a single reference CAD model. This approach is limited for Text-to-CAD as the prompt may specify certain design constraints or attributes but is inherently ambiguous and cannot determine all geometric and structural details required to uniquely define a CAD model. Geometric similarity metrics thus fail to account for the space of valid design variations within the degrees of freedom unconstrained by the prompt. Instead, in this work, Text-to-CAD is viewed as a code synthesis task where natural language descriptions is translated into an executable parametric program that produces valid 3D geometry. This perspective motivates an evaluation paradigm that moves beyond geometric comparison, instead leveraging the programmatic nature of CAD models as executable, testable code. We draw inspiration from the field of program synthesis where popular benchmarks such as HumanEval~\cite{Chen2021EvaluatingLL}, APPS~\cite{Hendrycks2021MeasuringCC}, SPoC~\cite{Kulal2019SPoCSP} and MBPP~\cite{Austin2021ProgramSW} evaluate generated code by executing test suites and a sample is correct only if it passes all tests.

\begin{figure}[t]
\setlength{\belowcaptionskip}{-0.5cm}
  \centering
  \includegraphics[width=\linewidth]{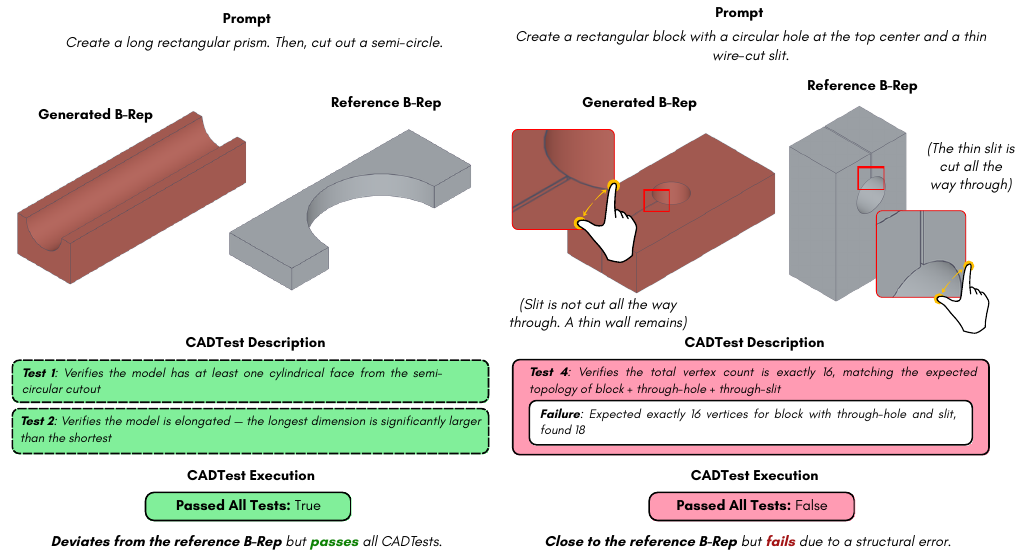}
  \caption{\textsc{CADTests} are executable programs that verify prompt specifications  directly on the generated geometry. \textsc{CADTests} can account for ambiguous design prompts \textit{(left)} and identify subtle structural errors \textit{(right)}. The figure ommits \textsc{CADTest} implementations for clarity.}
  \label{fig:teaser}
\end{figure}

In this work we introduce a novel \textit{test-driven} evaluation paradigm for Text-to-CAD based on \textsc{CADTests}. \textsc{CADTests} are executable software tests verifying whether a generated CAD model meets the design specifications of the input prompt. Concretely, each test is a Python code snippet that queries the generated CAD model through CadQuery~\cite{cadquery}, a Python library for programmatic CAD design that also exposes a rich API for geometric and topological analysis. CADQuery enables programmatic manipulation of the Boundary Representation (B-Rep), the standard data structure that describes a solid by its bounding surfaces, edges, and vertices, enabling fine-grained inspection of both the geometry and the topology of the generated model. It also provides expressive selectors that isolate specific geometric entities allowing tests to evaluate prompt-relevant parts of the geometry directly rather than reasoning over the whole model. \textsc{CADTests} use selectors along with B-Rep inspection primitives, including topology counts, bounding box dimensions, areas and volumes, center-of-mass queries, containment tests or line-to-geometry intersections that can be composed to verify complex geometric and topological CAD model properties.

Building on \textsc{CADTests}, we introduce \textsc{CADTestBench}, the first Text-to-CAD \textit{testing} benchmark, which pairs language prompts with suites of \textsc{CADTests} designed to evaluate prompt to CAD model correspondence. We develop an automated pipeline for \textsc{CADTests} synthesis and their iterative refinement, drawing on the established software engineering technique of Mutation Analysis~\cite{demillo1979program, 5487526, Papadakis2019ChapterS}. Mutated CAD models or simply \textit{mutants}, are CAD B-Reps that deliberately violate one or more geometric requirements specified in the prompt (e.g altering dimensions, removing features, or modifying topology) thereby acting as hard negative examples against which the discriminative power of the test suite is assessed. A test suite is first generated by a powerful Large Language Model (LLM), then validated against a ground-truth reference CAD model and a set of mutants. A refinement loop extends the suite with new tests and repairs existing ones until all tests pass on the reference while each mutant is rejected by at least one test.

Test-based evaluation offers several advantages. It is \textit{(1) reference-free}, assessing prompt compliance directly rather than via alignment with a ground-truth solution. It enables \textit{(2) explicit requirement verification}, treating every dimension and structural specification as a strict constraint. \textsc{CADTests} are also \textit{(3) interpretable}, with failing tests returning explicit log messages describing the cause of failure, and \textit{(4) efficient}, as they are fast, deterministic, and avoid the variability and inference costs of learnable semantic evaluators such as CLIP~\cite{Yuan2025CADEditorAL}  or VLLM-based methods~\cite{Yavartanoo2024Text2CADTT, Wang2025TexttoCADGT}. Fig. \ref{fig:teaser} demonstrates that the generated \textsc{CADTests} can account for valid design variations and capture subtle violations of prompt requirements.

\noindent
\textbf{Contributions}: The main contributions of this work are summarized as follows:
\begin{enumerate}
    \item We introduce \textsc{CADTests}, i.e executable software tests verify whether a generated CAD model satisfies the geometric and topological requirements specified by a natural language prompt. Based on \textsc{CADTests} we develop \textsc{CADTestBench}, the first \textit{test-based benchmark} for Text-to-CAD, constructed through an automated pipeline that iteratively refines test suites using mutation analysis.

    \item We conduct a comprehensive evaluation of recent Text-to-CAD methods on \textsc{CADTestBench}, providing insights into the progress and limitations of current approaches. We also demonstrate that \textsc{CADTests} can be leveraged as an effective feedback mechanism during generation, introducing test-based baselines that outperform existing Text-to-CAD methods.
    \item We validate the descriminative power of our \textsc{CADTests} via mutation analysis and a human study, demonstrating that test-based evaluation aligns closely with human judgment.

\end{enumerate}

\section{Related Work}

\noindent
\textbf{Text-Conditioned CAD Model Generation}: Generating CAD models from natural language descriptions has recently attracted significant interest~\cite{cadllama, Kolodiazhnyi2025cadrilleMC,khan2024text2cad,Yuan2025CADEditorAL,mallis2025cad,zhang2024flexcad,Badagabettu2024Query2CADGC,Idea2cad2025}. Text-to-CAD methods use \textit{text prompt - CAD model} pairs to train generative networks, with approaches ranging from fine-tuning of language backbones~\cite{khan2024text2cad, Li2024CADTA} to hierarchy-aware masking~\cite{zhang2024flexcad} and learning with direct preference optimization (DPO) through geometric reward~\cite{Guan2025CADCoderTG} or visual feedback~\cite{Kolodiazhnyi2025cadrilleMC, Wang2025TexttoCADGT}. A parallel \textit{training-free} direction leverages general-purpose foundational models to generate python CAD scripts directly. Strategies for enhancing generation performance include self-verification through question-answer generation~\cite{alrashedy2024generating}, execution feedback from an integrated CAD kernel~\cite{mallis2025cad}, geometric grounding from intermediate visual renderings~\cite{mallis2025cad, Idea2cad2025} and multi-agent schemes~\cite{Idea2cad2025}. Despite this growing body of work, the field lacks universally adopted benchmarks. Recent methods typically synthesize their own \textit{text prompt – CAD model} pairs on top of existing large-scale CAD datasets, primarily DeepCAD~\cite{wu2021deepcad}. In contrast, evaluation throughout the broader language and vision research space is increasingly based on curated benchmarks~\cite{Chen2021EvaluatingLL, Yue2023MMMUAM} rather than held-out subsets of training data. \textsc{CADTestBench} follows this practice, building on the recently proposed CADPrompt~\cite{alrashedy2024generating} text-to-CAD benchmark and extending it with executable \textsc{CADTests} that verify prompt requirements directly on the generated geometry.

\noindent
\textbf{Text-to-CAD Evaluation}: The most widely adopted evaluation metrics for Text-to-CAD are geometric similarity measures~\cite{khan2024text2cad,alrashedy2024generating,Li2024CADTA,Yuan2025CADEditorAL,Yavartanoo2024Text2CADTT,Wang2025TexttoCADGT,Guan2025CADCoderTG,Kolodiazhnyi2025cadrilleMC}, mainly bidirectional Chamfer Distance (CD) and Intersection over Union (IoU), computed against a ground-truth reference. Geometric similarity measures have been the standard for 3D reconstruction tasks~\cite{Rukhovich2024CADRecodeRE,Uy2021Point2CylRE}, where an input point cloud uniquely determines the target geometry. Text-to-CAD, however, is fundamentally ambiguous, and many geometrically distinct designs can correctly satisfy the same prompt, making geometric similarity to a single reference an ill-suited correctness criterion. Since CD, IoU can only be computed for generations that execute without errors, invalidity rates (IR) are reported separately, resulting in aggregated scores computed on method-specific subsets. Inspired by text-to-3D generation~\cite{Wu2024GPT4VisionIA}, recent methods have also employed automated \textit{visual inspection} via CLIP-based~\cite{Yuan2025CADEditorAL} or VLM scores on 2D renderings~\cite{Yavartanoo2024Text2CADTT, Wang2025TexttoCADGT}. Text-to-3D evaluation is often grounded in visual criteria~\cite{Wu2024GPT4VisionIA}, such as texture fidelity, geometric plausibility, and text-asset alignment, which can typically be judged from rendered 2D views. In contrast, CAD modelling is parametric and precision driven, where prompt requirements (e.g exact dimensions, solid connectivity, enclosed structures) must be verified explicitly on the generated geometry rather than visually assessed. This work addresses the limitations of existing metrics by directly verifying prompt compliance of the output geometry via \textsc{CADTests}.

\noindent
\textbf{Evaluation Protocols for Code Generation}: Modern evaluation of code generation relies on automated testing~\cite{Austin2021ProgramSW,Hendrycks2021MeasuringCC,Lachaux2020UnsupervisedTO,Chen2021EvaluatingLL,Li2022CompetitionlevelCG}, reporting metrics such as pass@\textit{k}~\cite{Chen2021EvaluatingLL,Kulal2019SPoCSP,Austin2021ProgramSW}, where $k$ code samples are generated per problem, and a generation is considered correct if any of the $k$ samples passes all tests. Test suites can be human-written~\cite{Hendrycks2021MeasuringCC,Kulal2019SPoCSP, Chen2021EvaluatingLL}, crowd-sourced~\cite{Austin2021ProgramSW} or derived from open source projects by collecting input output traces from integration tests~\cite{Chen2021EvaluatingLL}. Despite their widespread adoption, benchmark test suites are inherently insufficient to fully characterize correct program behavior, allowing incorrect implementations to pass all tests~\cite{Chen2021EvaluatingLL, Hendrycks2021MeasuringCC}. CodeContests~\cite{Li2022CompetitionlevelCG} mitigate this limitation to some extend by generating mutated inputs through small perturbations such as bit flips, integer offsets, or character substitutions.  In this work, we view Text-to-CAD as an instance of text-conditioned program synthesis and leverage test-driven evaluation principles from code generation for CAD model assessment, a perspective overlooked by prior work. Similarly to~\cite{Li2022CompetitionlevelCG}, we also employ mutated variants of CAD programs to construct test suites that capture prompt specifications, while remaining robust to the diversity of valid geometric outputs.

\section{Background}

\noindent
\textbf{Text-to-CAD Generation.} A CAD model is commonly represented using a Boundary Representation (B-rep), a data structure that describes a solid through its boundary surfaces, composed of faces, edges, and vertices together with their adjacency relationships~\cite{Lambourne2021BRepNetAT}. 
Formally, a B-rep $m \in \mathcal{M}$ with $\mathcal{M}$ denoting the set of B-reps, can be defined as $m = (\mathcal{F}, \mathcal{E}, \mathcal{V})$, where 
$\mathcal{F}$, $\mathcal{E}$, and $\mathcal{V}$ denote the sets of faces, edges, and vertices, respectively. A B-rep $m$ is obtained by executing a CAD program $f_p \in \mathbf{\Pi}$ such that
$m = \llbracket f_p \rrbracket$, where $\llbracket \cdot \rrbracket$ denotes the execution of a CAD program. This CAD program represents the sequence of modeling operations performed by a designer during the CAD creation process. In Text-to-CAD generation, existing methods aim to automatically synthesize such CAD programs from a textual prompt. The generated programs can take different forms, including sequences of parametric tokens defined by a Domain-Specific Language (DSL)~\cite{wu2021deepcad,khan2024text2cad,Yuan2025CADEditorAL}, or executable Python scripts built using CAD libraries such as CadQuery~\cite{alrashedy2024generating,Idea2cad2025,mallis2025cad,Guan2025CADCoderTG,Kolodiazhnyi2025cadrilleMC} or FreeCAD~\cite{mallis2025cad}. B-Rep models can be imported into CAD environments (\textit{eg.}, CadQuery) for fine-grained inspection of geometric and topological properties via the CadQuery API.

\noindent \textbf{Software Testing.} Code generation is often defined as the synthesis of standalone functions from text docstrings were evaluation is performed via execution of a comprehensive suite of \textit{automated tests}~\cite{Austin2021ProgramSW,Hendrycks2021MeasuringCC,Lachaux2020UnsupervisedTO,Chen2021EvaluatingLL,Li2022CompetitionlevelCG}. Text-to-CAD can be viewed as a specialised instance of the code generation paradigm, where the docstring corresponds to a design prompt and the generated CAD program produces a B-rep geometry as output that must fulfill well-defined requirements of the input prompt. In this work we employ \textit{property-based testing} to verify the generated B-rep models without relying on a single reference geometry. Property-based testing~\cite{MacIver2019} is a powerful technique~\cite{10.1145/1159789.1159792,7515466}, where correctness is verified by checking whether generated outputs satisfy a set of invariant properties rather than matching predefined input–output pairs. Recent work~\cite{Vikram2024CanLL} shows that property-based tests can be effectively generated by LLMs from docstring specifications. The same work also adopts \textit{mutation analysis}, a widely studied software testing technique~\cite{demillo1979program,5487526,Papadakis2019ChapterS} used to assess the fault detection capability of a test suite. In the Text-to-CAD context, a mutant is a faulty variant of a CAD program created by injecting a controlled artificial modification into its instructions, thereby producing a corrupted geometry. A test suite is said to \textit{kill} a mutant if at least one property test detects the injected fault, and the mutation score (Mscore) is defined as the fraction of killed mutants, providing a proxy measure of test coverage. In this work, we draw inspiration from property-based testing in software engineering and introduce \textsc{CADTests}, a set of CAD-specific property tests designed to evaluate Text-to-CAD outputs. We leverage mutation analysis both to measure the effectiveness of generated \textsc{CADTests} and to guide the design of more discriminative tests.

\section{\textsc{CADTests}}
\label{sec:method}

We define a \textsc{CADTest}, a property-based test $\mathcal{T}_i : \mathcal{M} \rightarrow \{0, 1\}$, as a boolean predicate evaluating a
generated CAD model $m \in \mathcal{M}$, indicating whether a specified geometric or structural constraint is satisfied.  Concretely, each $\mathcal{T}_i$ is implemented as a Python code snippet executed on the B-rep $m$. A test suite $\mathcal{T} = \{\mathcal{T}_1, \mathcal{T}_2, \ldots, \mathcal{T}_N\}$ consists of multiple such tests, each capturing a distinct property of the expected design. The overall evaluation is defined as the conjunction \(\mathcal{T}(m) = \bigwedge_{i=1}^{N} \mathcal{T}_i(m)\). The B-rep structure is well suited to this formulation: its topological and geometric entities can be 
directly queried to construct atomic tests, 
which can in turn be composed into higher-level tests encoding complex geometric and topological properties of the design.

\begin{figure}[t]
\setlength{\belowcaptionskip}{-0.6cm}
  \centering
  \includegraphics[width=0.98\linewidth]{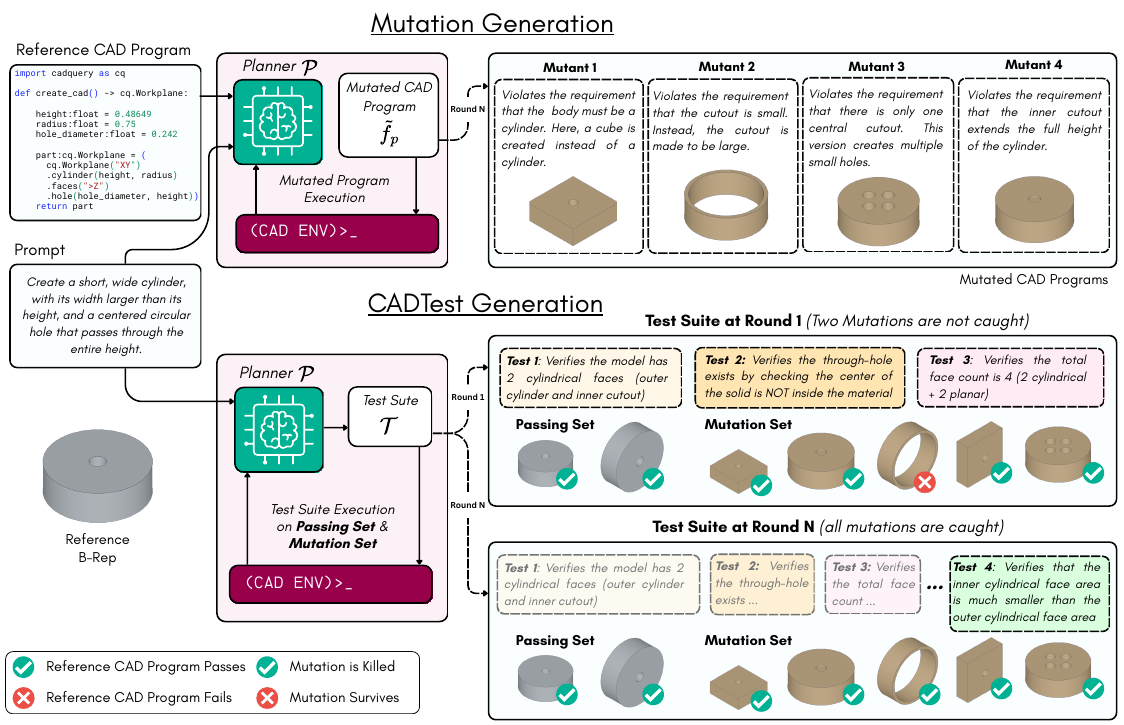}
  \caption{Illustration of the proposed \textit{CADTest} synthesis pipeline. \textit{(Top)} A design prompt and reference CAD program are provided to an LLM planner to generate CAD mutants. \textit{(Bottom)} The planner generates a test suite that is iteratively refined using execution feedback from the \textit{passing} and \textit{mutation} sets. In this example, a mutation that is not \textit{killed} in $R=1$ is detected in $R=N$ after adding a test comparing the volumes of the inner and outer cylindrical faces.}
  \label{fig:pipelineCADTests}
\end{figure}

\textsc{CADTests} are not readily available and must be synthesized. We propose 
an automatic \textsc{CADTests} synthesis pipeline operating on triplets $(p, f_p, m)$, where $p$ is a text prompt, 
$f_p$ is a reference CAD program satisfying $p$, and $m = \llbracket f_p \rrbracket$ is 
the corresponding B-rep model. The pipeline is orchestrated by an LLM planner $\mathcal{P}$, capable of advanced reasoning, which performs two main steps: (1) generating mutations of the reference CAD program, and (2) leveraging these mutations to refine \textsc{CADTests}. An overview of the pipeline is illustrated in Fig.~\ref{fig:pipelineCADTests}, and the different steps are described below.

\vspace{0.1cm}
\noindent\textbf{Mutation Generation.} The planner $\mathcal{P}$ is conditioned on 
both the text prompt $p$ and the reference solution $f_p$, and is tasked with generating 
a set of $K$ geometric mutants $\{\tilde{f}_p^1, \tilde{f}_p^2, \ldots, \tilde{f}_p^K\}$, 
where each mutant $\tilde{f}_p^k$ is a modified variant of $f_p$ that deliberately 
violates one or more geometric requirements specified in $p$. Each mutant is accompanied 
by a textual description $\tilde{d}_p^k$ disambiguating why the mutant fails to satisfy the design intent (eg. “violates the requirement that the cutout is centered”). To ensure program validity, each 
candidate mutant undergoes a ReAct~\cite{Yao2022ReActSR} refinement loop. The generated $\tilde{f}_p^k$ is executed in a CAD environment and, if a runtime 
error occurs, the trace is fed back to $\mathcal{P}$ as an observation, prompting it to reason about the cause of failure and emit a repaired program.

\vspace{0.1cm}
\noindent\textbf{\textsc{CADTest} Generation.} The planner $\mathcal{P}$ analyses the text prompt $p$ to generate a test suite $\mathcal{T}= \{\mathcal{T}_1, \mathcal{T}_2, \ldots, 
\mathcal{T}_N\}$ of initial \textsc{CADTests}, where $N$ is not fixed but determined by the 
number of verifiable properties the planner $\mathcal{P}$ identifies in $p$.
Each test $\mathcal{T}_i$ is accompanied by a textual description $d_i$ that clarifies the specific property of the prompt being verified. The planner's context is also augmented with a structured 
reference document detailing relevant CadQuery API methods, usage patterns, and a set 
of requirements that well-formed tests must satisfy. Note that generated tests are conditioned solely on the text prompt $p$, 
remaining agnostic to the reference solution $f_p$, ensuring 
that $\mathcal{T}$ captures the design intent of $p$ rather than the specifics of any particular implementation. The initial test suite might execute with runtime errors or provide insufficient coverage of prompt specifications. To refine generated \textsc{CADTests}, we introduce an \textit{iterative refinement pipeline} that improves $\mathcal{T}$ using feedback from a CAD execution environment. 

The generated tests are executed on both a \textit{passing set} of valid CAD models 
and a \textit{mutation set} composed of mutant CAD models that violate the prompt requirements. \textit{(i) Passing Set:} The passing set consists of the reference model 
$m = \llbracket f_p \rrbracket$ together with a collection of augmented variants obtained 
through similarity transformations,
$\{a(m) \mid a \in \mathcal{A}\}$. Each transformation $a \in \mathcal{A}$ is 
defined as $a = t \circ r \circ s$, where $s \in \mathbb{R}^+$ denotes a uniform 
scaling, $r \in SO(3)$ a 3D rotation, and $t \in \mathbb{R}^3$ a translation. 
Although the planner $\mathcal{P}$ never observes the reference program $f_p$, 
it may implicitly assume particular placements, orientations, or scales of the 
generated geometry. Applying similarity transformations therefore, encourages 
the synthesized tests to be invariant to the pose and scale of the 
evaluated B-rep $m$. \textit{(ii) Mutation Set:} The mutation set consists of the mutated models 
$\{\tilde{m}^k = \llbracket \tilde{f}_p^k \rrbracket\}$ and their augmentations  through similarity transformations 
$\{a(\tilde{m}^k) \mid a \in \mathcal{A}\}$. Since similarity transformations preserve 
geometric structure, any augmented mutant remains a violation of $p$, and augmenting 
the negative set in this way increases the number of mutants and thus the coverage of 
the refinement signal.

The test suite $\mathcal{T}$ is executed on all models from both sets, and the resulting diagnostic traces are collected. The planner $\mathcal{P}$ then analyzes these traces and, following a ReAct-style~\cite{Yao2022ReActSR} reasoning loop, produces an updated version of $\mathcal{T}$ by modifying existing tests, introducing new ones, or removing tests that are redundant or incorrect. Note that the planner is not conditioned on the reference CAD program $f_p$ nor on the mutated implementations themselves. Instead, it receives only a textual description of the applied mutation or augmentation parameters, together with diagnostic traces produced when executing the tests on the resulting geometries. Because each test returns an informative message in both success and failure cases, $\mathcal{P}$ receives rich feedback at every iteration, including failure explanations for reference models and false-negative explanations for mutants that remain undetected. The refinement process iterates until all tests pass on the passing set and every mutant in the mutation set is detected, or until a maximum of $R$ refinement rounds is reached. After $R$ rounds, any tests that still fail on the reference models are discarded from the final test suite $\mathcal{T}$.

\begin{table}[b]
\setlength{\tabcolsep}{3pt}
\centering
\vspace{-0.4cm}
\caption{\textsc{CADTests} generation performance for abstract prompts in terms of validity, soundness, and mutation score. We evaluate LLM planners and compare two prompting strategies: providing the planner with $20$ \textsc{CADTests}, or provide a \textsc{CADTest} authoring skill file.}
\resizebox{0.9\linewidth}{!}{%
\begin{tabular}{lcccccc}
\toprule
\multirow{2}{*}{\textbf{Model}} 
& \multicolumn{2}{c}{\textbf{Prompting}} 
& \textbf{Valid}\scriptsize(\#valid/\#total)
& \textbf{Sound}\scriptsize(\#sound/\#valid)
& \textbf{MScore} \\
& \textit{$20$-shot} & \textit{Skilled} & & & \\

\multirow{4}{*}{\small{\texttt{GPT-4.1}}}
& \xmark & \xmark & $54.0\%\text{\scriptsize(937/1733)}$ & $65.1\%\text{\scriptsize(610/937)}$ & 
$25.0\%$\\
& \xmark & \cmark & $86.6\%\text{\scriptsize(1466/1691)}$ & $71.5\%\text{\scriptsize(1049/1466)}$ & 
$51.1\%$\\
& \cmark & \xmark & $85.2\%\text{\scriptsize(1360/1597)}$ & $73.9\%\text{\scriptsize(1005/1360)}$ & 
$42.6\%$\\
& \cmark & \cmark & $90.5\%\text{\scriptsize(1570/1734)}$ & $75.9\%\text{\scriptsize(1191/1570)}$ & 
$48.4\%$\\

\multirow{1}{*}{\small{\texttt{GPT-5.2}}}
& \xmark & \cmark & $72.1\%\text{\scriptsize(1667/2312)}$ & $79.2\%\text{\scriptsize(1321/1667)}$ & 
$48.7\%$\\

\multirow{1}{*}{\small{\texttt{Claude-4.6-Sonnet}}}
& \xmark & \cmark & $97.8\%\text{\scriptsize (2543/2599)}$ & $\textbf{90.6}\%\text{\scriptsize(2303/2543)}$ & 
$63.0\%$\\

\multirow{1}{*}{\small{\texttt{Claude-4.6-Opus}}}
& \xmark & \cmark & $\textbf{98.3}\%\text{\scriptsize(2387/2428)}$ & $89.0 \%\text{\scriptsize(2125/2387)}$ & 
$\textbf{65.4}\%$\\

\bottomrule
\end{tabular}
}

\label{tab:cadtest_generation}
\end{table}

\begin{table}[t]
\caption{Effectiveness of the refinement loop. Scores are reported for the reference implementation, CAD mutants, as well as their augmented variants.}
\setlength{\tabcolsep}{4pt}
\centering
\resizebox{0.9\linewidth}{!}{%
\begin{tabular}{ccccccc}
\toprule
\small{\textbf{Prompt}}
& \small{\textbf{Model}}
& \small{\textbf{\#total}}
& \small{\textbf{Sound}}
& {\small\textbf{Sound}\(_{\text{Augm}}\)}
& \small{\textbf{MScore}}
& {\small \textbf{MScore}\(_{\text{Augm}}\)}\\

\small{\textit{Detailed}} & \multirow{2}{*}{\small{\texttt{Claude 4.6 Sonnet}}}
 & 3,091  &  0.982 & $0.89\pm0.02$ & 0.91 & $0.95\pm0.04$\\

\small{\textit{Abstract}} & 
& 2,989  & 0.972 & 0.92$\pm$0.01 & 0.93 & $0.89\pm$0.09\\
\bottomrule
\end{tabular}
}
\vspace{-0.1cm}
\label{tab:iterative_refinement}
\end{table}

\begin{table}[t]
\caption{Mutation score (MScore) \textit{before} and \textit{after} refinement for abstract and detailed prompts, measured on held-out mutants not seen during refinement.}
\setlength{\tabcolsep}{4pt}
\centering
\begin{tabular}{lcccc}
\toprule
& \multicolumn{2}{c}{\textbf{Abstract}} &
\multicolumn{2}{c}{\textbf{Detailed}}\\
\textbf{Pipeline Stage} & \textit{before-refinement}
& \textit{after-refinement}
& \textit{before-refinement}
& \textit{after-refinement}\\
\cmidrule{2-5}
\textbf{Mscore} & 0.552  &  0.730 & 0.641 & 0.787 \\
\bottomrule
\end{tabular}
\vspace{-0.1cm}
\label{tab:heldout_set}
\end{table}

\section{The \textsc{CADTestBench}}
\label{sec:cadtestbench}

We construct \textsc{CADTestBench} using the proposed iterative refinement pipeline. \textsc{CADTestBench} is derived from CADPrompt~\cite{alrashedy2024generating}, a Text-to-CAD benchmark of $200$ CAD programs, each paired with an \textit{abstract} prompt providing a high level design description and a \textit{detailed} prompt specifying additional geometric constraints such as exact dimensions, proportions, etc. Since tests verify prompt requirements, we generate separate test suites for each prompt type. In total \textsc{CADTestBench} comprises $5,937$ tests for both abstract and detailed prompt with an average of $\approx15$ tests per sample. Mutated CAD models used within our \textsc{CADTests} synthesis framework are generated using a \texttt{GPT-4.1} planner~\cite{openai2023gpt4}. We also generate a separate held-out set of mutants used only for evaluation via a different LLM (\texttt{Claude-Sonnet-4.6}). Additional statistical analysis of the generated \textsc{CADTests} and the mutants produced by our synthesis pipeline is provided in the supplementary material. To generate \textsc{CADTests}, we use a \texttt{Claude-Sonnet-4.6} planner~\cite{anthropic2025claude4}. The refinement pipeline iterates for $R=4$ rounds. For both the reference and mutated CAD models, we sample $5$ augmentations by applying small scaling, translating, and rotating the model by 90 degrees along different axes. For the \textit{detailed} partition where prompts specify exact dimensions, we omit uniform scaling. Tests are organized using two complementary classification schemes. First, they are grouped by the \textit{prompt requirement} they verify, as multiple tests may evaluate the same requirement through complementary checks. Second, tests are categorized by the type of property they assess. We define the following categories: \textit{(1) solid/shell validity}, \textit{(2) topology checks} (e.g., number of faces or edges), \textit{(3) geometry type checks}, \textit{(4) dimensional and ratio checks}, \textit{(5) volumetric checks}, and \textit{(6) spatial relationship checks} such as alignment, symmetry, and offsets. A detailed description of categories and dataset statistics is provided in the supplementary.

\textbf{Evaluation}. We report 4 complementary metrics. The \textit{Pass-Rate (PR)} measures the percentage of samples where \textit{all} \textsc{CADTests} pass indicating full prompt compliance. The \textit{Requirement Score (RS)} evaluates each prompt requirement group independently and measures the fraction of prompt requirements satisfied per sample, where a requirement is considered satisfied only if all tests in its group pass. We also report \textit{Accuracy (Acc)} for different \texttt{CADTest} property types and the \textit{Invalid-Ratio (IR)}, which captures the proportion of generated samples that execute with runtime errors. Note that
we count invalid generations as failures for all metrics, in contrast to prior protocols (e.g IoU, CD) that discard them and thus evaluate each method on its own subset.

\begin{table}[t]
\setlength{\tabcolsep}{5pt}
\centering
\caption{Evaluation of Text-to-CAD methods on \textsc{CADTestBench}. Results are reported for both the abstract and detailed partitions. Text-to-CAD baselines are evaluated using different LLM planners. Performance is measured in terms of pass rate, requirement score, and invalidity rate.}
\resizebox{0.95\linewidth}{!}{%
\begin{tabular}{lccccccccc}
\toprule
 \textbf{Method}&\textbf{Planner} & \textbf{IR}~$\downarrow$ & \textbf{RS}~$\uparrow$ & \textbf{PR}~$\uparrow$ &  \textbf{IR}~$\downarrow$ & \textbf{RS}~$\uparrow$ & \textbf{PR}~$\uparrow$\\
  &  & \multicolumn{3}{c}{\textit{Detailed}} & \multicolumn{3}{c}{\textit{Abstract}} \\

\cmidrule{3-8}
\multicolumn{8}{l}{\textit{\underline{Finetuned Method}}}\\ 
 \small{Text2CAD}~\cite{khan2024text2cad} & - & 0.005 & 0.405&  0.025 & 0.015 & 0.603 & 0.085 \\

\multicolumn{8}{l}{\textit{\underline{Train-Free Methods}}}\\ 
\small{CadCodeVerify}~\cite{alrashedy2024generating} & \texttt{GPT-4.1} & 0.025 & 0.794 &  0.410 & 0.040 & 0.880 & 0.630\\

\small{10-shots} & \texttt{GPT-5.2} & 0.175 & 0.718 & 0.465& 0.210& 0.742 & 0.580 \\
\small{Skilled} &  \texttt{GPT-5.2} & 0.190 & 0.688 & 0.430& 0.180 & 0.766 & 0.600 \\
\small{ReAct}& \texttt{DeepSeek-V3} & 0.035&  0.752 &  0.370 & 0.005 & 0.821& 0.390\\
\small{ReAct}& \texttt{Qwen3-Coder-480B} & 0.025 & 0.793 &  0.425 & 0.01 & 0.881& 0.605\\
 \small{ReAct}& \texttt{GPT-5.2}& 0 & 0.835 &0.480 & 0.005 & 0.916 & 0.695 \\
\small{ReAct + Image} & \texttt{GPT-5.2}& 0.025 &0.822 & 0.485 & 0.005 & 0.916 & 0.695 \\

\small{10-shots}  & \texttt{Claude-4.6-Sonnet} &  0.155& 0.744&  0.510 &  0.115&  0.830 & 0.64\\ 
\small{Skilled}& \texttt{Claude-4.6-Sonnet} & 0.185 &0.727& 0.505&  0.130 & 0.810& 0.640\\
\small{ReAct}  & \texttt{Claude-4.6-Sonnet}& 0 & 0.874& 0.580 & 0 &  0.929& 0.715\\

\multicolumn{8}{l}{\textit{\underline{Test-based Baselines}}}\\ 
\small{CADTests} & \texttt{GPT-5.2}& 0.020 & 0.824 &  0.500 & 0.015 & 0.925 & 0.72 \\
\small{CADTests + Log} & \texttt{GPT-5.2} &  0.015 & 0.852 & 0.515 & 0.020 &  0.922 &  0.700 \\
\small{CADTests}& \texttt{Claude-4.6-Sonnet} & 0.005& 0.882 & 0.590&   0 &  0.953&  0.765 \\
\small{CADTests + Log} & \texttt{Claude-4.6-Sonnet}& 0& \textbf{0.897} & \textbf{0.625} &  0 & \textbf{0.962}& \textbf{0.810}\\
 
 \bottomrule

\end{tabular}
 }
\label{tab:results}
\end{table}

\begin{figure}[t]
\setlength{\belowcaptionskip}{-0.7cm}
  \centering
\includegraphics[width=\linewidth]{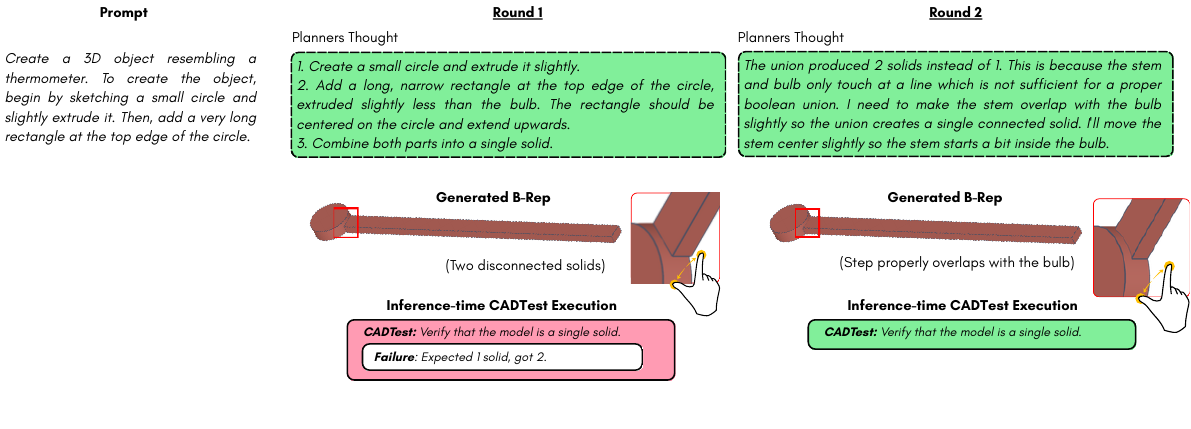}
    \vspace{-0.8cm}
  \caption{Generation trajectory for the \textit{(CADTest)} baseline. During ReAct style inference the planner produces both the CAD program and a set of tests. \textsc{CADTest} execution logs revealing that the boolean union resulted in two separate solids. The planner then adjusts the design so the parts overlap. }
  \label{fig:testingeneration}
  \vspace{0.3cm}
\end{figure}

\section{Experiments}

\noindent
\textbf{\textsc{CADTest} Evaluation.} To assess the effectiveness of LLM planners in generating \textsc{CADTests}, we adopt an evaluation methodology similar to \cite{Vikram2024CanLL}. We report a \textit{validity ratio} (\textbf{Valid}) for generated \textsc{CADTests} along with a \textit{soundness ratio} (\textbf{Sound}) that is the percentage of tests that pass on a reference CAD program and \textit{mutation score} (\textbf{MScore}) that is the percentage of CAD mutants detected by the test suite. The MScore measures the fraction of injected prompt violations detected by the test suite, serving as a proxy for test coverage over the space of mutated programs. Note that, this experiment evaluates the ability of the planner to generate tests without being provided with a CAD environment or being exposed to the passing and mutated sets. Results for abstract prompts are reported in Tab.~\ref{tab:cadtest_generation}. Models from the Claude family generate valid and sound tests, achieving the highest mutant coverage by detecting approximately $\approx65\%$ of CAD mutants. Still, a large fraction of mutants remains uncaught, motivating the need for iterative refinement of \textsc{CADTests}. Based on this evaluation, we use \texttt{Claude-4.6-Sonnet} to generate \textsc{CADTestBench}, as it offers strong performance at a lower cost than Opus.

Tab.~\ref{tab:iterative_refinement} reports the effectiveness of the refinement pipeline, which improves the generated test suites using execution feedback from the passing and mutant sets. Scores are reported for the reference solutions, the CAD mutants, and their augmented variants. In this experiment, the same CAD mutants serve both as feedback for the refinement loop and as the basis for the mutation score. After $4$ refinement rounds, the MScore exceeds $90\%$ for both detailed and abstract prompts, showing that the pipeline captures previously undetected mutants. Validity and soundness are reported for completeness but only valid and sound \textsc{CADTests} are retained in \textsc{CADTestBench}. To assess whether the refined \textsc{CADTests} generalize beyond the mutants seen during refinement, we also measure MScore on a held-out mutant set generated independently by a different model (\texttt{Claude-4.6-Sonnet}), which is never used during refinement. As shown in Tab.~\ref{tab:heldout_set}, refinement improves the held-out mutation score by $+17.80\%$ for abstract prompts and $+14.61\%$ for detailed prompts. Since these mutants were never seen by the refinement loop, the gains indicate that refinement produces tests that catch unseen mutants rather than overfitting to refinement mutation set.

\begin{figure}[t]
\setlength{\belowcaptionskip}{-0.3cm}
  \centering
\includegraphics[width=0.98\linewidth]{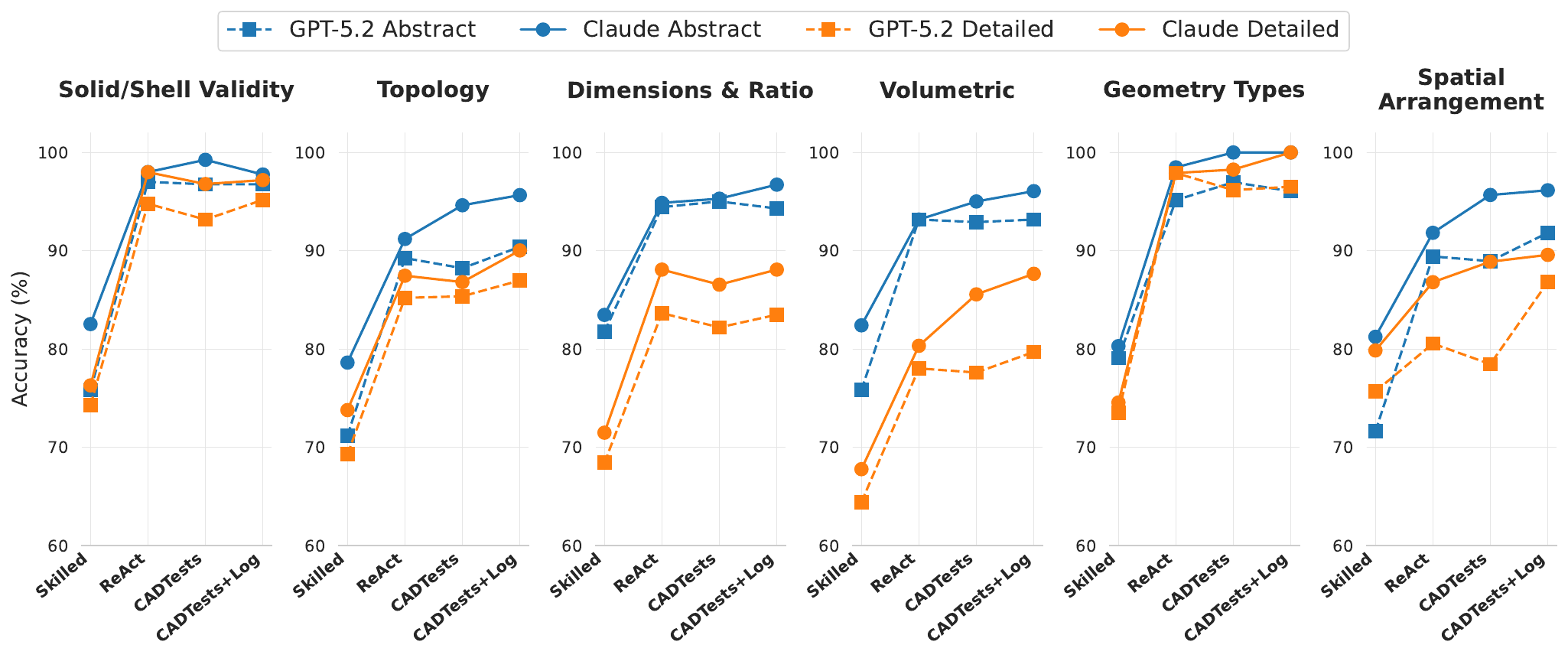}
  \vspace{-0.3cm}
  \caption{Category-level evaluation of Text-to-CAD baselines. Results are reported in terms percentage of tests passed on for category.}
  \label{fig:categories}
\end{figure}

\noindent
\textbf{Text-to-CAD Evaluation}. Using \textsc{CADTestBench}, we conduct a thorough evaluation of Text-to-CAD methods. We compare multiple \textit{training-free} baselines based on general-purpose VLMs. Specifically we use: \textit{(K-shot)} an LLM planner conditioned on \textit{k} demonstrations, each consisting of a prompt paired with its corresponding CADQuery program, \textit{(Skilled)} using a markdown skill file providing guidelines for authoring \textit{CADQuery} code, \textit{(ReAct)} iteratively executing generated code and feeding runtime errors back to the planner, \textit{(ReAct + Image)} additionally provides four-view renderings of the generated model and the \textit{(CADCodeVerify)} framework of~\cite{alrashedy2024generating}. For methods trained on synthetic data, we consider Text2CAD~\cite{khan2024text2cad} and CADRille~\cite{Kolodiazhnyi2025cadrilleMC}. A detailed description of all examined Text-to-CAD methods is provided in the supplementary material.

We also introduce two simple baselines that integrate \textsc{CADTests} into the generation loop. \textit{(CADTest)} conditions the planner to generate both CAD code and \textsc{CADTests}, which are executed in a ReAct loop for iterative refinement. The generated code and tests are executed within a ReAct loop, where \textsc{CADTest} execution feedback is returned to the planner and used to iteratively refine both the CAD
program and the associated \textsc{CADTests}. An illustration of the effect of test-based feedback during generation is shown in Fig.~\ref{fig:testingeneration}. \textit{(CADTest + Log)} extends the CADTest baseline by injecting log queries (\texttt{print} statements) that expose intermediate geometric states during execution. These logs provide the planner with quantitative feedback on the evolving CAD geometry, enabling it to verify whether each stage of the generation process behaved as intended.

The bulk of our results are provided in Tab. ~\ref{tab:results} from which several insights emerge.

\begin{enumerate}[topsep=1pt]
    \item[(i)]\textit{Performance on Detailed vs Abstract Prompts}. A consistent observation is that all baselines achieve higher performance on abstract prompts than on detailed prompts when evaluated with \textsc{CADTests}. This difference arises because detailed prompts impose specific requirements, making them inherently more difficult to satisfy than abstract prompts.
    
    \item[(ii)]\textit{Limitations of Synthetic Finetuning}. Text2CAD~\cite{khan2024text2cad} underperforms compared to training-free baselines. CADRille~\cite{Kolodiazhnyi2025cadrilleMC} resulted in degenerate solutions, and we excluded it from reported results. Although fine-tuned methods can demonstrate strong performance on held-out test sets, they fail to generalize on \textsc{CADTestBench}. Their training corpora consist of synthetically generated prompts paired with CAD models restricted to the sketch-and-extrude paradigm, and cannot capture prompts and geometries of \textsc{CADTestBench}.
    \item[(iii)] \textit{Effect of VLM Planners on Performance.} We observe that using \texttt{Claude} as the planner yields a significant performance increase over \texttt{GPT}-based variants across all methods, highlighting the choice of the underlying foundation model as a key driver of performance.
    \item[(iv)] \textit{Integration with the CAD Kernel.} Baselines that incorporate ReAct style CAD kernel integration achieve higher performance and much lower invalidity rates. Execution feedback allows the planner to iteratively correct runtime errors and produce valid CAD programs.
    \item[(v)] \textit{Visual Feedback.} Visual feedback via multi-view renderings does not improve performance. We limited further exploration since visual feedback significantly increases token usage.
    \item[(vi)] \textit{CADTests Enhance Text-to-CAD Performance.} The \textit{(CADTests)} baselines constitute the most effective feedback mechanism, yielding a consistent performance increase across all metrics. For instance, \textit{(CADTests+Log)} achieves an approximately 10\% improvement in pass rate and 3\% in requirement score on abstract prompts. This highlights the potential of \textsc{CADTests} as an effective mechanism incorporating CAD kernel feedback into CAD code generation.

\end{enumerate}

\begin{table}[b]
\setlength{\tabcolsep}{4pt}
\caption{Pass-rate agreement with human consensus (five annotators, majority vote, Fleiss’ $\kappa = 0.53$, $N = 125$) for \textsc{CADTests} and the LVM-based metric of~\cite{Wang2025TexttoCADGT}.}
\centering

\begin{tabular}{lcccc}
\toprule
\textbf{Metric} & \textit{Accuracy} & \textit{Precision} & \textit{Recall} & \textit{F1} \\
\cmidrule{2-5}
LVM~\cite{Wang2025TexttoCADGT} & 0.472  & 0.912  & 0.357  & 0.515  \\
\textsc{CADTests} (PR Score) & \textbf{0.864}  & \textbf{0.945}  & \textbf{0.878} & \textbf{0.910} \\
\bottomrule
\end{tabular}%

\label{tab:human_alignment_pass}
\end{table}

\begin{table}[t]
\setlength{\tabcolsep}{4pt}
\caption{ROC-AUC of continuous score metrics against human consensus labels (five annotators, majority vote, $N = 125$)}
\centering
\begin{tabular}{lcccc}
\toprule
 \textbf{Metric} & Chamfer Distance & CLIP Score~\cite{Yuan2025CADEditorAL} & LVM~\cite{Wang2025TexttoCADGT} & \textsc{CADTests} (RS Score)\\
\cmidrule{2-5}
\textbf{ROC-AUC}  & 0.674 & 0.598 & 0.680 & \textbf{0.848} \\
\bottomrule
\end{tabular}
\label{tab:human_alignment_auc}
\end{table}

In Fig. \ref{fig:categories}, we report the accuracy per category across \textsc{CADTest} property type categories using two LLM planners (GPT and Claude) on both abstract and detailed prompts. The results show that solid validity and geometric types are the least challenging categories, while topological, dimensional, volumetric, and spatial requirements are more difficult to satisfy. This suggests that current LLM planners still have room for improvement in handling precise unit-based, topological, and spatial reasoning constraints.


\begin{figure}[t]
\setlength{\belowcaptionskip}{-0.5cm}
  \centering
    \vspace{-0.3cm}
  \includegraphics[width=0.95\linewidth]{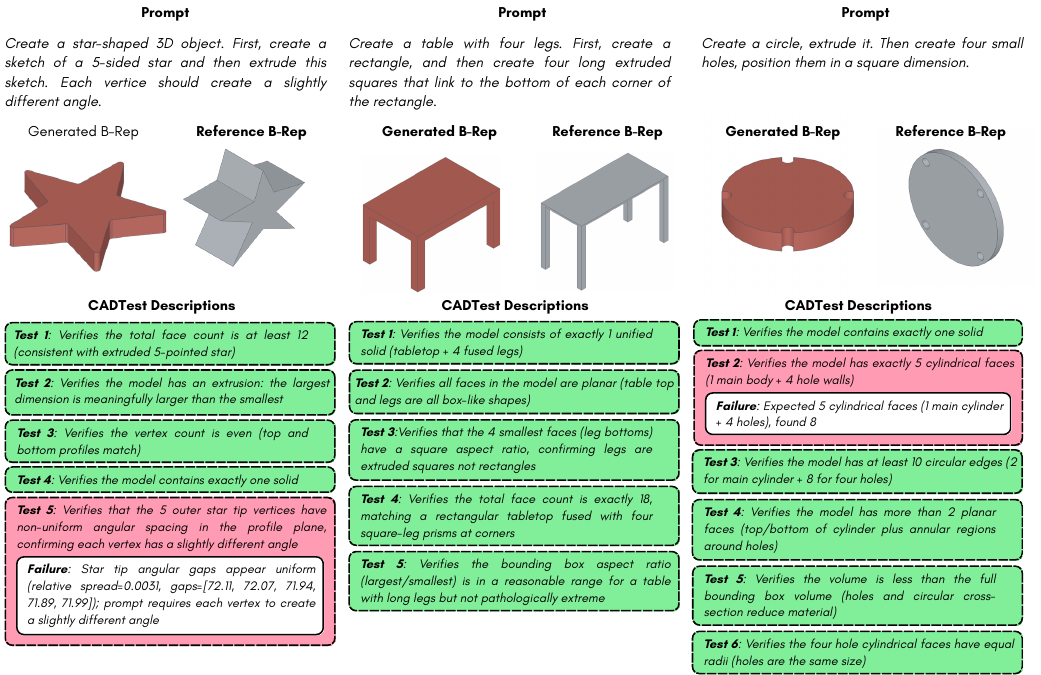}
  \vspace{-0.2cm}
  \caption{Example generations from the (\textit{CADTest}) baseline evaluated via \textsc{CADTests}. For clarity, only a subset of test descriptions is shown and full test implementations are omitted. }
  \label{fig:testexamples}
\end{figure}

\noindent
\textbf{Evaluation of \textsc{CADTestBench}}. To assess how well \textsc{CADTestBench} aligns with human judgment, we conduct a human study. Five expert annotators are presented with $125$ generation results sampled at random from our evaluated baselines and asked to judge whether each generation matches the design prompt (full protocol in the supplementary material). Consensus is formed by majority vote. The annotators achieve moderate inter-annotator agreement (Fleiss’s $\kappa = 0.53$), reflecting the inherent ambiguity of judging prompt–CAD correspondence. We measure agreement between this consensus, \textsc{CADTests} and commonly used Text-to-CAD evaluation metrics. Pass-rate agreement is reported in Tab.~\ref{tab:human_alignment_pass} and ROC-AUC for continuous scores in Tab.~\ref{tab:human_alignment_auc}. We observe that \textsc{CADTests}-based evaluation aligns closer to human judgments than all other metrics considered. Fig. \ref{fig:testexamples} shows qualitative results for Text-to-CAD generation and test execution results. In Fig. \ref{fig:testexamples}~\textit{(left)}, \textsc{CADTests}
effectively verify that the generated star has uniform angular
spacing. In Fig. \ref{fig:testexamples}~\textit{(middle)}, the generated model exhibits geometric variations with respect to the reference
that are not constrained by the prompt and are therefore not penalized by \textsc{CADTests}. In Fig. \ref{fig:testexamples}~\textit{(right)}, tests detect that the holes are not enclosed but instead form open slots that break through the disk boundary.

\section{Conclusions}
\label{sec:conclusions}
In this work, we introduce \textsc{CADTests}, executable software
tests that verify whether a generated CAD model satisfies the
specifications of an input prompt, and propose \textsc{CADTestBench}, the
first \textit{test-driven} benchmark for Text-to-CAD. We develop an automatic \textsc{CADTest} synthesis pipeline achieving high discriminative power via iterative refinement and also demonstrate that \textsc{CADTest} can serve as an effective feedback
mechanism during generation, yielding baselines that outperform existing Text-to-CAD methods. \textit{Limitations:} Despite automated testing being the prevalent evaluation paradigm for code generation, misclassifications do occur by exploiting limitations in test coverage~\cite{Chen2021EvaluatingLL, Hendrycks2021MeasuringCC}. \textsc{CADTestBench} inherits this limitation. The proposed \textsc{CADTest} synthesis pipeline mitigates this and achieves high mutation scores, but the discriminative power of any test-suite is finite, and \textsc{CADTest} verification may diverge from human judgment.

\section{Acknowledgments}
This work is supported by the National Research Fund (FNR), Luxembourg, under the BRIDGES2021/IS/16849599/FREE-3D project and the JUMP2025/19750191/Morfis project. 

\bibliographystyle{plainnat}
\bibliography{main}

\maketitlesupplementary

\appendix

\setcounter{page}{1}
\renewcommand{\thesection}{\Alph{section}}
\definecolor{light-gray}{gray}{0.95}
\newcommand{\code}[1]{\colorbox{light-gray}{\texttt{\scriptsize#1}}}

This supplementary material includes additional details and illustrations that were not included on the main.

\section{\textsc{CADTestBench} Analysis}

\textsc{CADTestBench} contains 200 samples, with each sample comprising a pair of language prompts along with a suite of \textsc{CADTests} designed to evaluate prompt-to-CAD-model correspondence. \textsc{CADTestBench} is based on the CADPrompt dataset that is dominated by sketch-and-extrude operations and combinations of 3D high-level primitives (e.g., boxes, cylinders). CADPrompt also includes models with more advanced operations such as fillet, shell, and sweep. Concretely, 23\% of models use only sketch-and-extrude, 53\% only 3D primitives, 17.5\% a combination of both, and 6.5\% involve other operations (fillet, shell, sweep).

\begin{figure}[H]
  \centering
  \includegraphics[width=\linewidth]{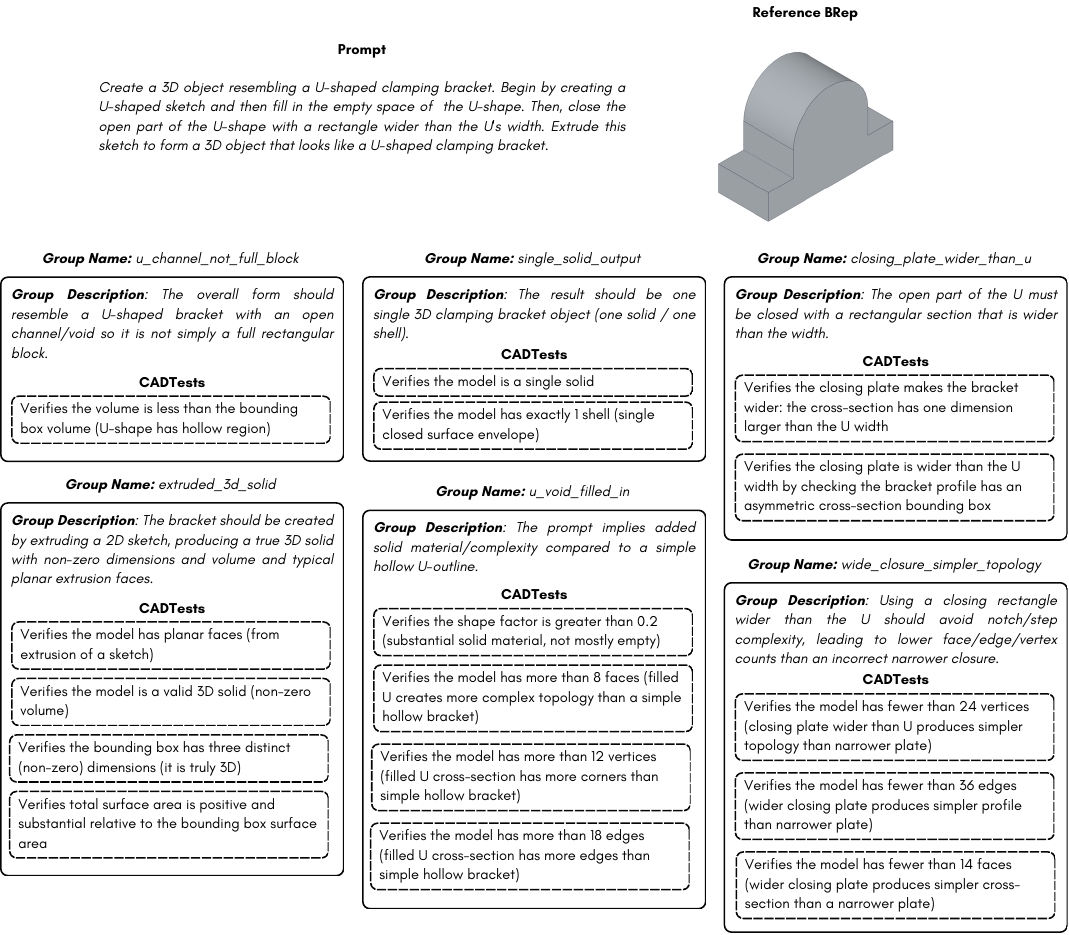}
  \caption{Examples of prompt requirement groups for a \textsc{CADTestBench} sample. The figure shows the name and description of each of the six identified groups, along with the \textsc{CADTest} assigned to each group. For clarity of visualization we only show the description of \textsc{CADTests} and omit their implementation.}
  \label{fig:groups}
\end{figure}

\begin{figure}[H]
  \centering
  \includegraphics[width=\linewidth]{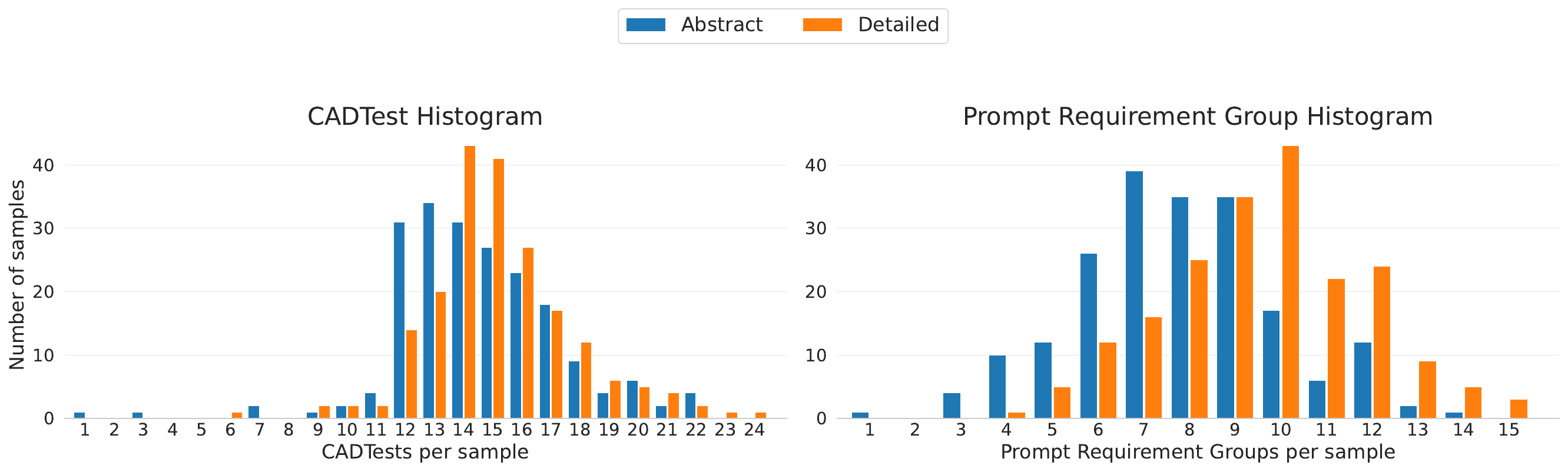}
  \caption{Distribution of \textit{(left)} \textsc{CADTests} counts and \textit{(right)} discovered prompt requirement groups per benchmark sample for abstract and detailed prompts.}
  \label{fig:histograms}
\end{figure}

\begin{figure}[H]
\centering
\begin{subfigure}[c]{0.45\textwidth}
    \centering
    \setlength{\tabcolsep}{3pt}
    \begin{tabular}{lcc}
     \multicolumn{3}{c}{\textbf{Benchmark Counts}}\\
    \toprule
     & \textit{Detailed} & \textit{Abstract} \\
    \multicolumn{3}{l}{\underline{\textsc{CADTests}}}\\
     Total & 3,037 & 2,9000 \\
     Avg & 15.2 & 14.5 \\
     \\
     \multicolumn{3}{l}{\underline{\textsc{Prompt Requirement Groups}}}\\
     Total & 1,920 & 1,554 \\
     Avg & 9.6 & 7.8 \\
    \midrule
    \end{tabular}

\end{subfigure}%
\hfill
\begin{subfigure}[c]{0.5\textwidth}
    \centering
    \includegraphics[width=\textwidth]{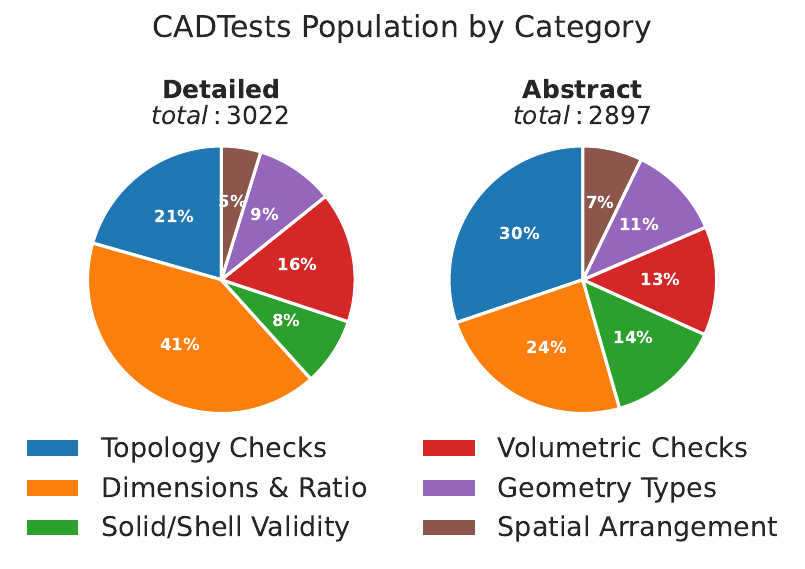}

\end{subfigure}
  \caption{\textit{(left)} Counts of generated CADTests and as well as prompt requirement groups across the detailed and abstract partitions of the \textsc{CADTestBench}. \textit{(right)} Share of \textsc{CADTests} by category for the detailed and abstract prompts.}
\label{tab:counts}
\end{figure}

Prompts for each sample are provided at two levels of description as \textit{detailed} prompts and \textit{abstract} prompts. Through our proposed iterative refinement pipeline, we construct a test suite comprising 5,937 tests. \textsc{CADTests} for each sample are organized into \textit{prompt requirement groups}, i.e groups of \textsc{CADTests} that verify the same prompt requirement. Prompt requirement groups are assigned by a \texttt{GPT-4.1} classifier. Examples of prompt requirement groups for a \textsc{CADTestBench} sample are shown in Fig. \ref{fig:groups}. The number of \textsc{CADTests} and prompt requirement groups per prompt type is reported in Fig. \ref{tab:counts} \textit{(left)}. In Fig. \ref{fig:histograms}, we present histograms illustrating the distribution of CADTests and prompt requirement groups generated per sample. Both Fig. \ref{tab:counts}~\textit{(left)} and Fig. \ref{fig:histograms} show that detailed prompts lead to a larger number of generated \textsc{CADTests}, since the additional information introduces more requirements from which \textsc{CADTests} can be synthesized.

Generated \textsc{CADTests} are also organized using a complementary classification scheme based on the type of property they evaluate. \textsc{CADTest} category classification is performed using a \texttt{GPT-4.1} classifier. \textsc{CADTests} are classified into the following categories:

\begin{itemize}

\item \textbf{Solid \& Shell Validity}  
This category captures checks that verify whether a model forms a valid, well-defined solid. These tests ensure properties such as the presence of a single solid body, a single connected shell, positive non-zero volume, non-degenerate 3D extent, and the absence of disconnected components. It also includes constraints on the number of solids in the model (e.g., exactly one solid, at least one solid, or a bounded range of solids).

\item \textbf{Topology Checks}  
This category includes tests that evaluate the structural complexity of a model by examining its topological elements. Typical checks involve counts of faces, edges, and vertices, either as exact values or bounded constraints. The category also includes comparisons against simple baselines, such as requiring more faces than a basic primitive, as well as typed counts when the goal is to verify quantities of specific elements (e.g., cylindrical faces or circular edges). 

\item \textbf{Geometric Types}  
This category covers tests that verify the presence or absence of specific geometric element types in the model. Examples include checks for planar faces, cylindrical faces, circular edges, or straight edges, without requiring exact counts. 

\item \textbf{Dimensions \& Ratios}  
This category includes tests that verify geometric measurements and proportional relationships. These checks involve absolute bounding box dimensions, aspect ratios, equal cross-section measurements, identification of the largest or smallest axis, individual face areas, feature diameters derived from edges, wall thickness values, and other dimension-based constraints.

\item \textbf{Volumetric Checks}  
This category captures tests that assess properties related to the total occupied volume of the model. These include comparisons between measured and expected volume, volume derived from dimensions, volume-to-bounding-box shape factors, and fill factors.

\item \textbf{Spatial Arrangement}  
This category includes tests that evaluate the spatial relationships between features or between features and the overall object. Examples include center-of-mass alignment with the bounding box center, concentric or coaxial alignment, offsets from edges, feature placement (e.g., centered, corner-based, or asymmetric), symmetry checks, arrangement patterns, through-hole versus blind pocket classification, and geometric relationships such as parallelism or perpendicularity.

\end{itemize}

\begin{figure}[H]
  \centering
  \includegraphics[width=\linewidth]{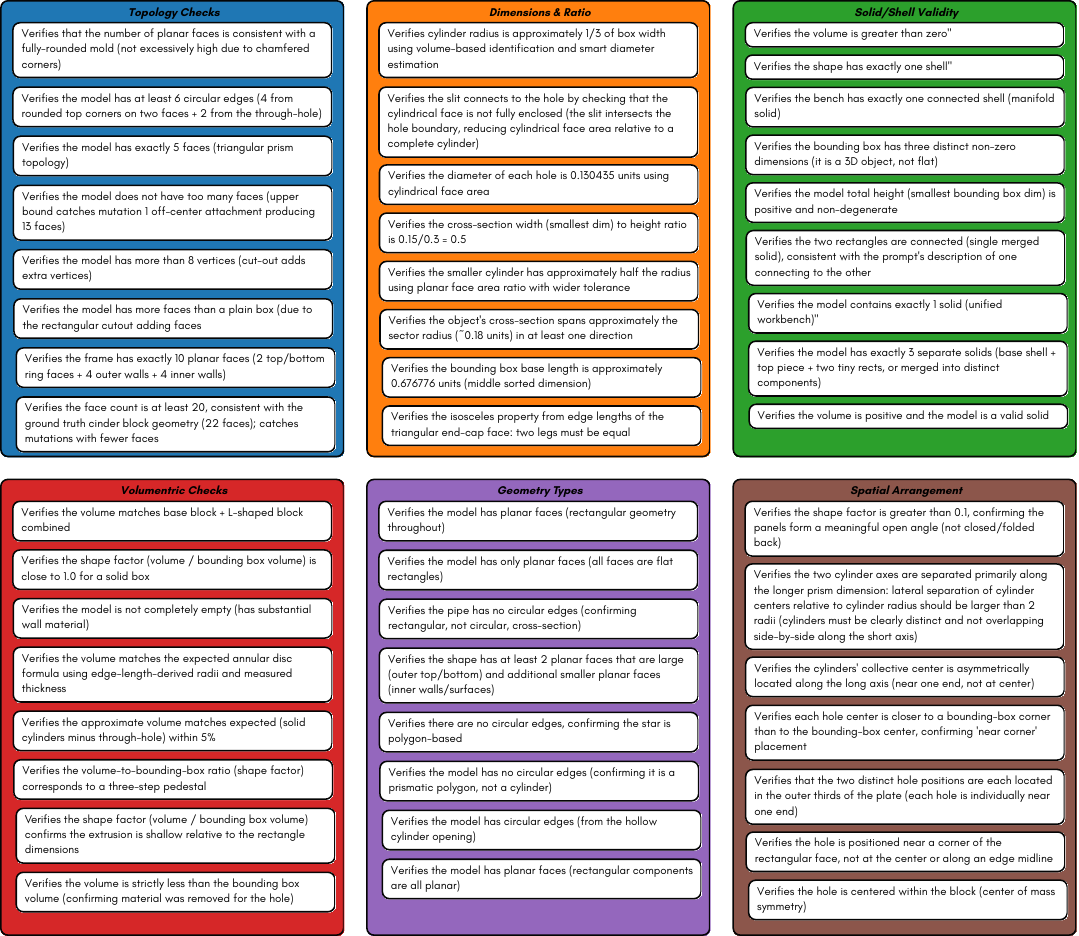}
  \caption{Example \textsc{CADTest} descriptions from multiple \textsc{CADTestBench} samples, grouped according to the six \textsc{CADTest} categories.}
  \label{fig:testcategoriesexamples}
\end{figure}

In Fig. \ref{tab:counts}~\textit{(right)}, we present a pie chart showing the distribution of \textsc{CADTest} categories for the detailed and abstract partitions. The detailed partition contains a higher proportion of \textit{Dimensions and Ratios} checks, reflecting the presence of explicit dimensional requirements in the detailed prompts. Examples of \textsc{CADTest} descriptions from \textsc{CADTestBench} assigned to the aformentioned categories are shown in Fig. \ref{fig:testcategoriesexamples}.

\begin{figure}[H]
  \centering
    \vspace{-0.4cm}
  \includegraphics[width=\linewidth]{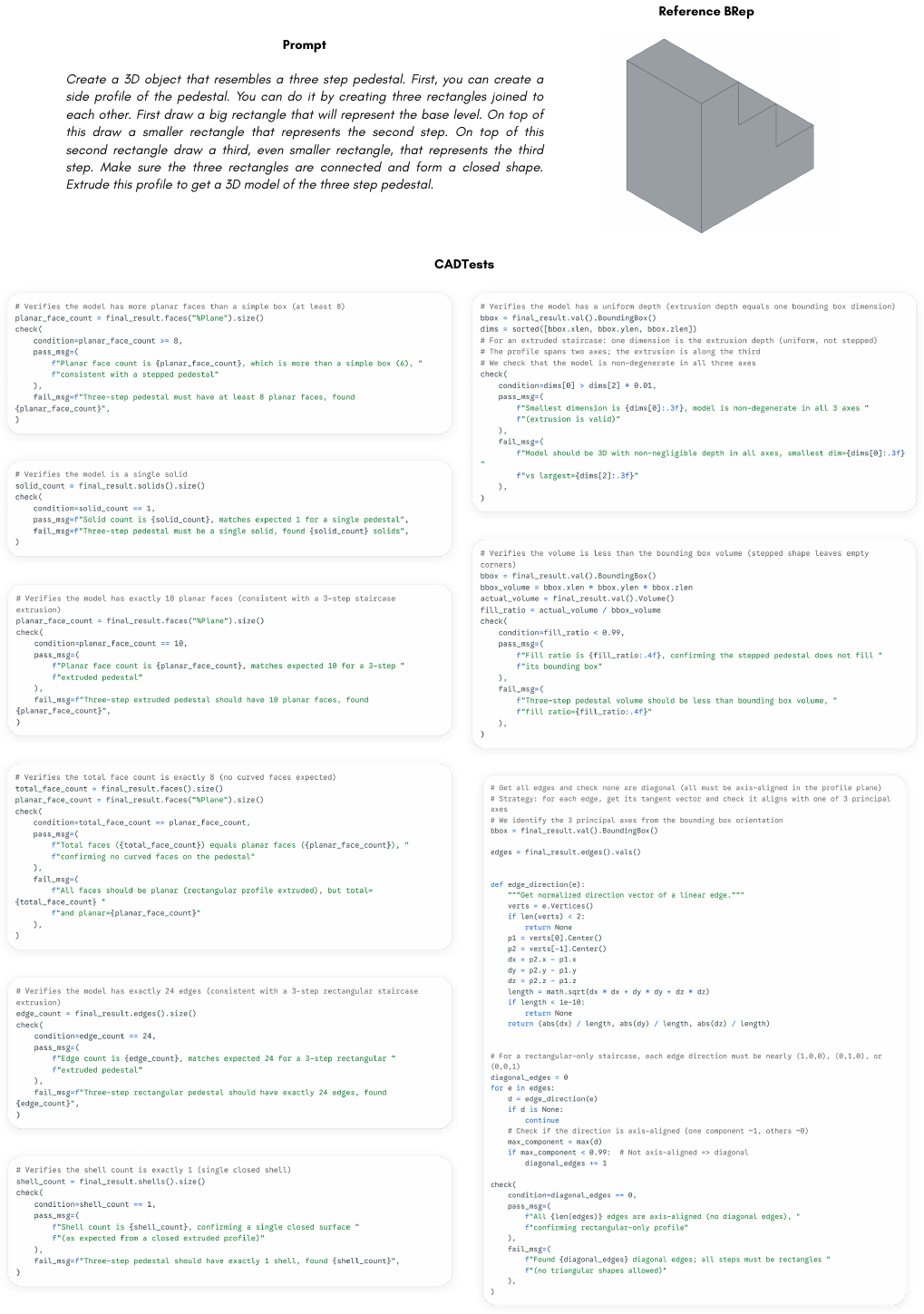}
  \caption{Example implementations of generated \textsc{CADTests} for a \textsc{CADTestBench} sample using the CADQuery API.}
  \label{fig:cadtestcode}
\end{figure}

\begin{figure}[H]
  \centering
  \includegraphics[width=0.7\linewidth]{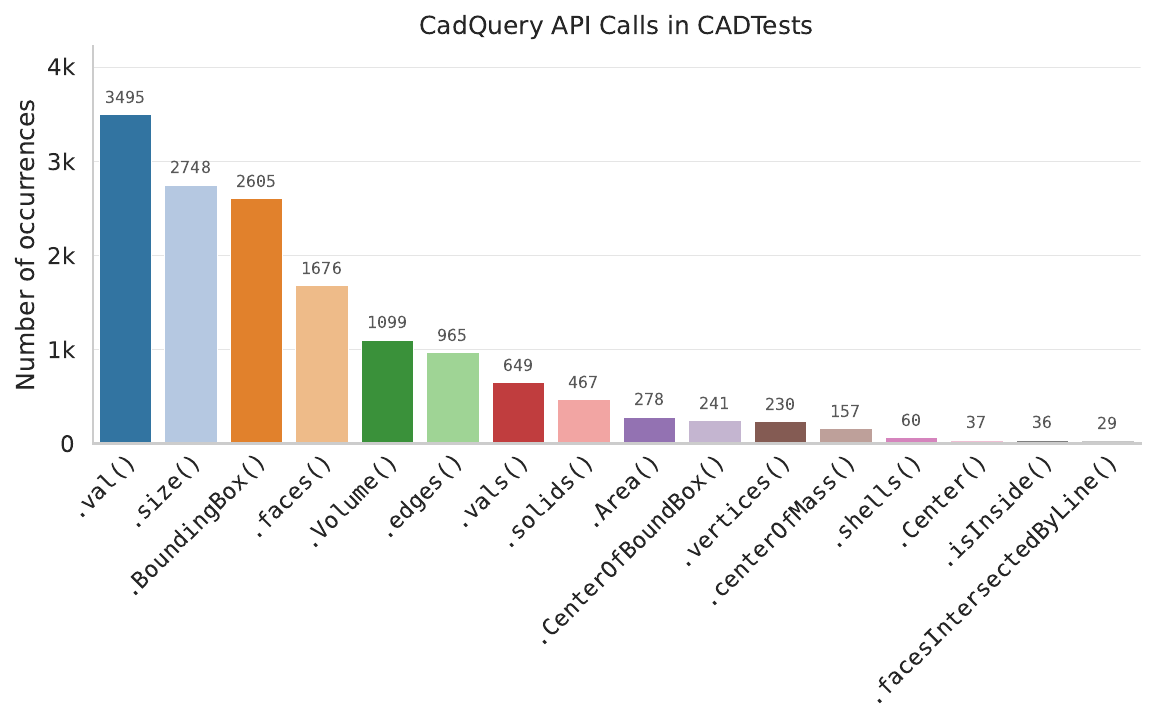}
  \caption{ API call frequency across \textsc{CADTests}, showing how often each CADQuery API method appears in test code. The method \texttt{val()} is the most frequently used, as it retrieves the underlying object from the workplane and is required in all tests to access the geometry being analyzed.}
  \label{fig:cadquery}
\end{figure}

The generated \textsc{CADTests} rely on the CADQuery API to perform topological and geometric analysis. Fig. \ref{fig:cadquery} shows the frequency of different CADQuery API calls across the generated \textsc{CADTests}. An example including \textsc{CADTests} code snippets for the generated test suite is shown in Fig. \ref{fig:cadtestcode}.

\section{Discussion}

\noindent
\textbf{Numerical tolerances.} Numerical tolerances are essential for reliable geometric testing. \textsc{CADTests} use tolerance-aware comparisons for operations such as position, distance, containment, and intersection checks. For example, verifying that a hole is centered requires comparing its position with the face center within a numerical tolerance rather than using exact equality. To reduce sensitivity to floating-point errors and object scale, the planner is explicitly instructed to avoid fixed absolute tolerances and instead define them relative to measured geometric quantities, such as the 2D or 3D bounding-box extent. This is particularly important for abstract prompts, where the object scale is unconstrained. Relative tolerances make the tests scale-aware and more robust to minor numerical discrepancies. This requirement is included in the \textsc{CADTest} guidelines supplied to the planner.

\noindent
\textbf{Evaluating generated B-Reps via \textsc{CADTests}}. \textsc{CADTests} are applied directly to the boundary representation (B-Rep), the standard parametric CAD data structure that describes a solid by its bounding surfaces, edges, and vertices. Operating on the B-Rep lets \textsc{CADTests} inspect both the geometry and the topology of a model. Because they consume B-Reps rather than scripts, \textsc{CADTests} also apply to methods such as Text2CAD, which predict token sequences in the DeepCAD representation rather than CAD Python programs: we simply compile the sequence and evaluate the resulting B-Rep. This also means \textsc{CADTests} are entirely independent of any step-by-step feature history. They operate on the final solid, never on the program or construction sequence that produced it. A \textsc{CADTest} inspects only the geometric and topological properties of the result, not how it was built.

\textbf{\textsc{CADTests} on implicit surfaces}. For implicit surfaces and meshes, a subset of our tests still applies, namely those that depend only on sampled or global geometry: bounding-box dimensions, surface areas and volumes, center-of-mass queries, containment tests, and line-to-geometry intersections. However, because these representations do not encode parametric primitives, the queries they support are less expressive. A triangular mesh, for instance, approximates a cylinder with many planar facets rather than the single cylindrical face a B-Rep provides, so topology counts (e.g., "exactly two cylindrical faces"), geometry-type checks, and inspection of geometric parameters become unreliable or undefined.

\clearpage

\section{CAD Mutants}
\begin{figure}[H]
  \centering
  \includegraphics[width=\linewidth]{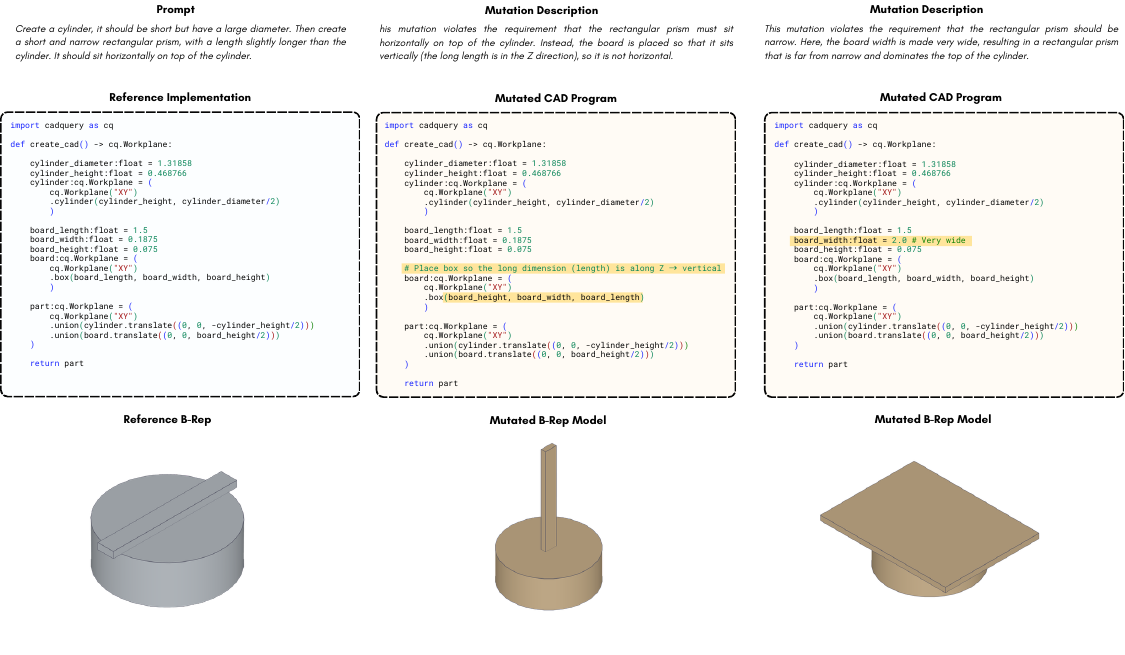}
  \caption{Depiction of CAD mutants. Our mutation generation pipeline takes a design prompt and a CAD program reference implementation and produces a mutated CAD program, along with a mutation description that explains why the mutant fails to satisfy the design intent.}
  \label{fig:mutated_programs}
\end{figure}

\begin{figure}[H]
  \centering
  \includegraphics[width=\linewidth]{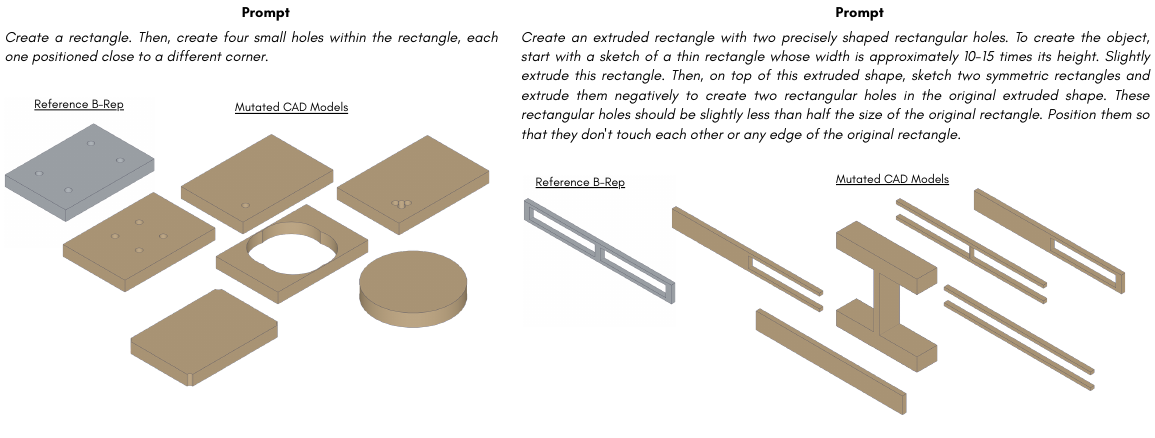}
  \caption{Examples of \textsc{CADTestBench} samples together with generated mutated CAD models that deliberately violate the requirements specified in the input prompt. The figure showcases only mutated CAD BReps and omits mutated CAD programs and the corresponding mutation descriptions for clarity.}
  \label{fig:mutatedbreps}
\end{figure}

Each CAD mutant is a modified variant of a CAD program that deliberately violates one or more prompt requirements. Mutants are generated via our mutation generation pipeline and further filtered via IoU computation w.r.t the reference CAD program to ensure that the synthesized mutation measurably modifies the generated shape.  Examples of mutated CAD programs are depicted in Fig. \ref{fig:mutated_programs}. More examples of mutated CAD BReps are shown in Fig. \ref{fig:mutatedbreps}.

\section{Text-to-CAD Methods and Baselines}

Using \textsc{CADTestBench}, we conduct a thorough evaluation of several baselines and recent methods for Text-to-CAD generation. We provide a detailed overview of the examined methods, grouped into three categories: methods trained on synthetic Text-to-CAD datasets, training-free approaches, and proposed baselines leveraging \textsc{CADTests} to enhance Text-to-CAD generation.

\vspace{0.1cm}
\noindent
\textbf{Finetuned Methods}. We consider Text2CAD~\cite{khan2024text2cad}, a transformer-based model trained on synthetic Text-to-CAD data derived from DeepCAD~\cite{wu2021deepcad}, to generate parametric token sequences from natural language descriptions. We also evaluated the multimodal CADRille~\cite{Kolodiazhnyi2025cadrilleMC}, trained on highly detailed prompts from the Text2CAD dataset~\cite{khan2024text2cad}. In our experiments, CADRille produced degenerate solutions on \textsc{CADTestBench}, and we excluded it from reported results.

\vspace{0.1cm}
\noindent
\textbf{Train-Free Methods}. We evaluate a set of training-free baselines that leverage general-purpose vision language models (VLMs) for Text-to-CAD generation.

\begin{itemize}

\item \textit{K-shot}.  
An LLM based planner for Text-to-CAD generation conditioned on 10 demonstrations. Each demonstration consists of a prompt paired with its corresponding CADQuery generated program, allowing the model to imitate the structure and style of valid CADQuery implementations.

\item \textit{Skilled}.  
The planner is conditioned on a skill document written in markdown that provides guidelines for writing CADQuery code. The document describes common CADQuery patterns, modeling strategies, and best practices to guide code generation.

\item \textit{ReAct}.  
This method extends the skilled planner with a ReAct loop~\cite{Yao2022ReActSR}. Generated CAD code is executed in a CAD environment and runtime errors are returned to the planner as observations. The planner then iteratively regenerates the CAD program to correct the detected issues.

\item \textit{ReAct + Vision}.  
An extension of the ReAct baseline that incorporates visual feedback. Four view renderings of the generated CAD model are provided to the model, enabling visual grounding of the generated geometry during the code generation process.

\item \textit{CADCodeVerify~\cite{alrashedy2024generating}}.  
A framework that iteratively improves CAD code using visual validation. A vision language model generates validation questions about renderings of the produced models, identifies inconsistencies or errors, and uses this feedback to guide subsequent code corrections. In this work CADCodeVerify was combined with a \texttt{GPT-4} planner as in~\cite{alrashedy2024generating}.

\end{itemize}

\vspace{0.1cm}
\noindent
\textbf{Test-Based Baselines}. We introduce two baselines that integrate testing into the generation pipeline, allowing the model to verify and refine generated CAD programs using execution feedback.

\begin{itemize}

\item \textit{CADTest}.  
This baseline introduces a CADTest generation step immediately after CAD code generation. The planner is conditioned to produce both the CAD program and a set of tests used to evaluate it. The generated code and tests are executed within a ReAct loop, where execution feedback is returned to the planner and used to iteratively refine both the CAD program and the associated tests. 

\item \textit{CADTest + Log}.  
This baseline extends \textit{CADTest} by injecting log messages that query intermediate geometric and topological properties of the generated model. These logs expose the intermediate state of the CAD geometry to the planner, enabling it to access additional information from the environment and verify whether each stage of the generation process behaves as intended.

\end{itemize}

\begin{figure}[H]
  \centering
  \includegraphics[width=\linewidth]{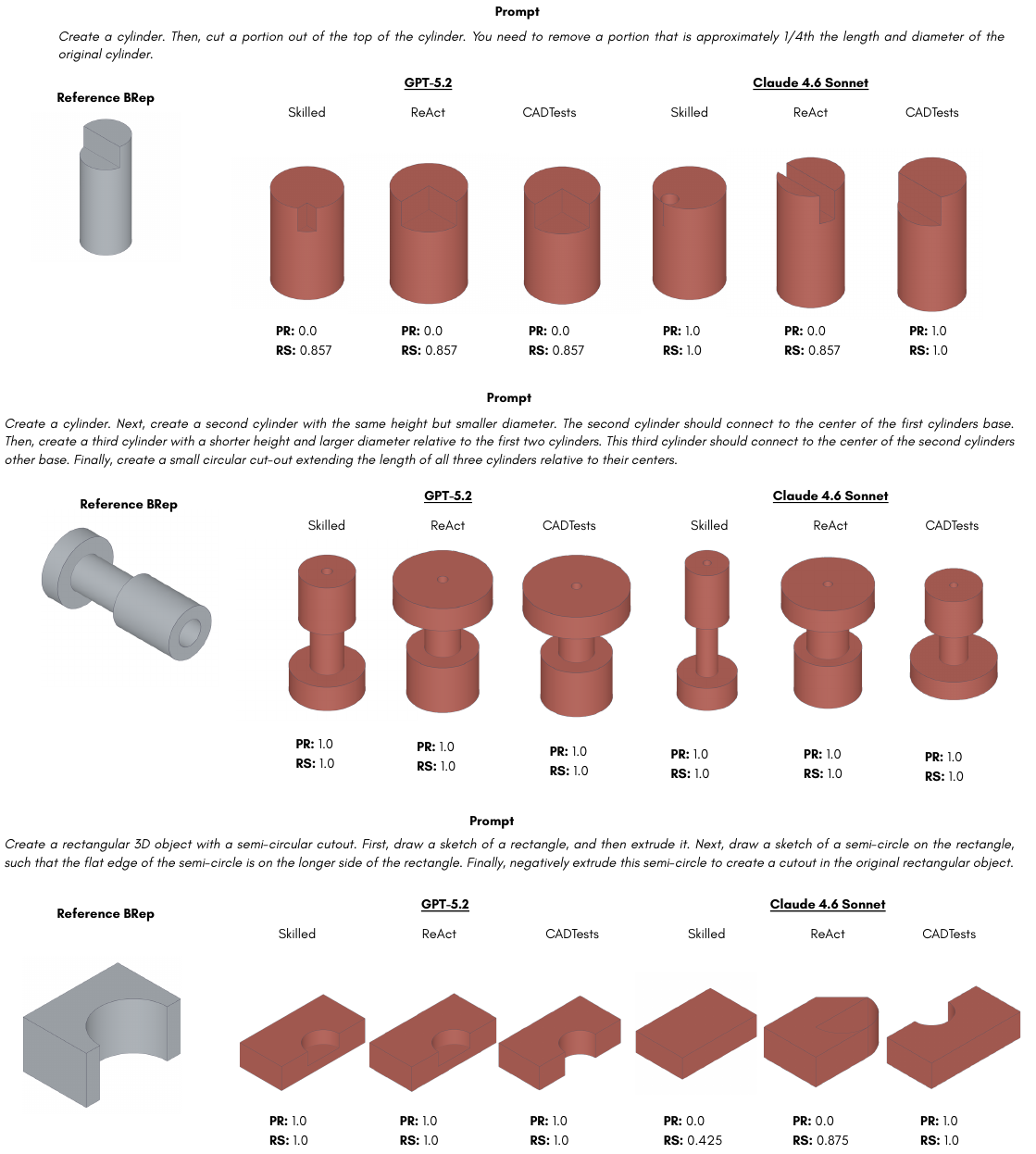}
  \caption{Qualitative evaluation results for the \textit{Skilled}, \textit{ReAct}, and \textit{CADTest} baselines using \texttt{GPT-5.2} and \texttt{Claude 4.6 Sonnet} planners. The figure shows the generated BRep for each baseline together with the score assigned by the \textsc{CADTest} suite. We report the pass rate (PR), which equals 1 when all tests are passed, and the requirement score (RS), defined as the fraction of prompt requirements satisfied for each sample.}
  \label{fig:qualitativeevalresults}
\end{figure}

In Fig. \ref{fig:qualitativeevalresults} we include qualitative results of evaluation scores assigned by \textsc{CADTestBench} on Text-to-CAD generation predictions generated by various baselines.

\noindent
\textbf{Model versions.} Tab.~\ref{tab:model_ids} lists the exact model snapshot used for each LLM planner. Anthropic's dateless identifiers map to a single fixed snapshot.

\begin{table}[h]
\centering
\caption{Model identifiers used in all experiments.}
\begin{tabular}{ll}
\toprule
\textbf{Model} & \textbf{Identifier} \\
\midrule
\texttt{GPT-5.2} & \texttt{gpt-5.2-2025-12-11} \\
\texttt{GPT-4.1} & \texttt{gpt-4.1-2025-04-14} \\
\texttt{Claude-4.6-Sonnet} & \texttt{claude-sonnet-4-6} \\
\texttt{DeepSeek-V3} & \texttt{DeepSeek-V3-0324} \\
\texttt{Qwen3-Coder-480B} & \texttt{Qwen3-Coder-480B-A35B-Instruct} \\
\bottomrule
\end{tabular}
\label{tab:model_ids}
\end{table}

\noindent
\textbf{Decoding.} We use temperature $0$ for all generation as well as evaluation, except for mutant generation, where we use $0.7$ to increase mutant diversity.

\vspace{0.1cm}
\noindent
\textbf{Run-to-Run Variability.} Repeating every experiment across all models is cost-prohibitive, so we instead run a targeted repeatability analysis on one baseline. We repeat the \textit{Skilled} \texttt{GPT-5.2} experiment on the detailed partition 4 times, keeping the benchmark, prompts, and evaluation protocol unchanged. Tab.~\ref{tab:variability} compares the value reported in the main paper with the mean and standard deviation over all four runs.

\begin{table}[h]
\centering
\caption{Run-to-run variability of \textit{Skilled} \texttt{GPT-5.2} on the detailed partition (4 runs). All values in \%.}
\vspace{0.3cm}
\begin{tabular}{lcc}
\toprule
\textbf{Metric} & \textbf{Reported} & \textbf{Mean $\pm$ SD (4 runs)} \\
\midrule
IR~$\downarrow$ & 19.0 & 18.63 $\pm$ 0.48 \\
RS~$\uparrow$ & 68.8 & 68.95 $\pm$ 0.29 \\
PR~$\uparrow$ & 43.0 & 42.75 $\pm$ 0.65 \\
\bottomrule
\end{tabular}
\label{tab:variability}
\end{table}

The reported values closely match the multi-run means, and all standard deviations are at most $0.65$ percentage points. We attribute this low variability partly to decoding at temperature $0$, although commercial APIs do not guarantee deterministic outputs even under this setting.

\noindent
\textbf{Inference cost.} Tab.~\ref{tab:cost} reports the estimated API cost of evaluating each main baseline on \textsc{CADTestBench} with the \texttt{GPT-5.2} planner. Iterative baselines are more expensive, since each refinement round adds execution feedback to the context. \textsc{CADTests}-based generation adds a moderate overhead over ReAct, which buys the largest performance gains in Tab.~\ref{tab:cost}.

\begin{table}[h]
\centering
\caption{Estimated API cost (USD) of evaluating each baseline on \textsc{CADTestBench} with the \texttt{GPT-5.2} planner.}
\begin{tabular}{lc}
\toprule
\textbf{Baseline} & \textbf{Cost (\$)} \\
\midrule
10-shots & 3.52 \\
Skilled & 5.98 \\
ReAct & 25.40 \\
\textsc{CADTests} & 30.20 \\
\textsc{CADTests} + Log & 36.23 \\
\bottomrule
\end{tabular}
\label{tab:cost}
\end{table}

\section{Human Evaluation of \textsc{CADTestBench}}
To assess how well CADTestBench aligns with human judgement, we conduct a human evaluation study with 5 expert annotators. The annotators are postgraduate researchers with experience in computer vision and CAD related research. We fully specify the protocol below.

\noindent
\textbf{Annotation instructions.} Each annotator is shown a design prompt and a generated 3D CAD model and answers a single binary question: does the generated model satisfy the requirements stated in the prompt? Annotators judge the model only against what the prompt explicitly specifies. Prompts are often incomplete, and any design choice the prompt leaves unconstrained is treated as free. A model satisfies the prompt if and only if it meets every stated requirement.

\noindent
\textbf{Interface.} Annotators are given the prompt text and the generated CAD model as a STEP file, which they inspect in CAD software (FreeCAD).

\noindent
\textbf{Blinding.} Annotators were blind to the CADTests output, to which baseline generated each sample, and to any reference solution.

\noindent
\textbf{Class balance.} Samples are drawn at random from our evaluated baselines. Human consensus over the five annotators labeled 78.4\% (98/125) as satisfying the prompt and 21.6\% (27/125) as not. Because the classes are imbalanced, we report precision, recall, F1, and AUC rather than accuracy alone.

\clearpage

\clearpage


\end{document}